\documentclass[10pt,twocolumn,letterpaper]{article}

\usepackage{iccv}
\usepackage{times}
\usepackage{epsfig}
\usepackage{graphicx}
\usepackage{amsmath}
\usepackage{amssymb}

\usepackage{subfig}
\usepackage{booktabs}
\usepackage{multirow}
\usepackage{longtable}
\usepackage{tabularx}
\usepackage{ifthen}

\newboolean{arxiv}
\setboolean{arxiv}{true}

\usepackage[pagebackref=true,breaklinks=true,letterpaper=true,colorlinks,bookmarks=false]{hyperref}

\ifthenelse{\boolean{arxiv}}{
\iccvfinalcopy % *** Uncomment this line for the final submission
}{ }

\usepackage{amsthm}
\usepackage{amsmath}
\usepackage{amssymb}
\usepackage{color}
\usepackage{xspace}
\usepackage{dsfont}

\def\srez{GAN}

\newcommand{\mo}[1]{}
\newcommand{\jon}[1]{}
\newcommand{\ryan}[1]{}
\newcommand{\todo}[1]{}

\newcommand{\comment}[1]{}

\newcommand{\logsumexp}[0]{\mathrm lse}

\renewcommand{\vec}[1]{\boldsymbol{\mathbf{#1}}}

\def\ourname{pixel recursive\xspace}
\def\Ourname{Pixel recursive\xspace}
\newcommand{\cond}[0]{A}
\newcommand{\prior}[0]{B}

\def\btheta{\vec{\theta}}
\def\bthetah{\widehat{{\btheta}}}
\def\thetah{\hat{{\theta}}}

\def\bw{\vec{w}}
\def\bx{\vec{x}}
\def\by{\vec{y}}
\def\y{y}
\def\ys{\y^{*}}

\def\Real{\mathbb{R}}

\def\modelp{{p}_{\theta}} % model prob
\def\modelph{{p}_{\thetah}} % model prob
\DeclareMathOperator{\softmax}{softmax}

\def\bys{{\mathbf y}^*}
\def\dataset{\mathcal{D}}

\def\int{\mathrm{int}}

\renewcommand{\th}[1]{{#1}^{\text{th}}}

\newcommand{\trans}[1]{{#1}^{\ensuremath{\mathsf{T}}}}

\def\eg{{\em e.g.,}}
\def\ie{{\em i.e.,}}

\def\etal{{\em et al.}}
\def\vs{{\em vs.}\xspace}

\newcommand{\figref}[1]{Figure~\ref{#1}}

\newcommand{\1}[1]{\mathds{1}[#1]}

\def\supp{Supp. material}
\def\NN{Nearest N.}
\def\regression{ResNet $L_2$}
\def\boldregression{\textbf{ResNet} $\mathbf{L_2}$}

\def\iccvPaperID{2837} % *** Enter the ICCV Paper ID here

\ificcvfinal\fi
\begin{document}

%%%%%%%%% TITLE
\title{Pixel Recursive Super Resolution}

\author{
Ryan Dahl
\thanks{Work done as a member of the Google Brain Residency program (g.co/brainresidency).}
\qquad
Mohammad Norouzi 
\qquad
Jonathon Shlens  \\
 Google Brain \\ {\tt\small \{rld,mnorouzi,shlens\}@google.com} 
}%

\maketitle
%\thispagestyle{empty}

%%%%%%%%% ABSTRACT
\begin{abstract}
Super resolution is the problem of artificially enlarging a low
resolution photograph to recover a plausible high resolution
version. In the regime of high magnification factors, the problem is
dramatically underspecified and many plausible, high resolution images
may match a given low resolution image.  In particular, traditional
super resolution techniques fail in this regime due to the
multimodality of the problem and strong prior information that must be
imposed on image synthesis to produce plausible high resolution
images. In this work we propose a new probabilistic deep network
architecture, a {\it pixel recursive super resolution model}, that is
an extension of PixelCNNs to address this problem. We demonstrate that
this model produces a diversity of plausible high resolution images at
large magnification factors.  Furthermore, in human evaluation studies
we demonstrate how previous methods fail to fool human
observers. However, high resolution images sampled from this
probabilistic deep network do fool a naive human observer a
significant fraction of the time.
\end{abstract}

%%%%%%%%% BODY TEXT
\section{Introduction}

The problem of {\em super resolution} entails artificially enlarging a
low resolution photograph to recover a corresponding plausible image
with higher resolution \cite{survey}. When a small magnification is
desired (\eg~$2\times$), super resolution techniques achieve
satisfactory results~\cite{sun2008image,
  Fattal:2007:IUV:1276377.1276496, huang1999statistics, shan2008fast,
  kim2010single} by building statistical prior models of
images~\cite{FoE, Aharon, mixtures} that capture low-level
characteristics of natural images.

\begin{figure}
\begin{center}
\begin{tabular}{@{\hspace{.2cm}}c@{\hspace{.2cm}}c@{\hspace{.2cm}}c@{\hspace{.2cm}}}
\multicolumn{3}{c}{}\\
$8\!\times\!8$ input & $32\!\times\!32$ samples & ground truth\\[.1cm]
% /cns/od-d/home/rld/super_resolution/exp102/A_one_hot/samples/
{\includegraphics[width=.29\linewidth]{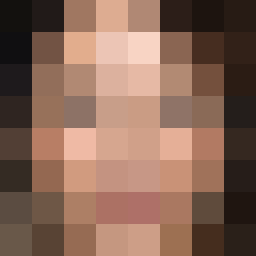}} &
{\includegraphics[width=.29\linewidth]{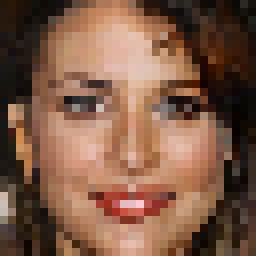}} &
{\includegraphics[width=.29\linewidth]{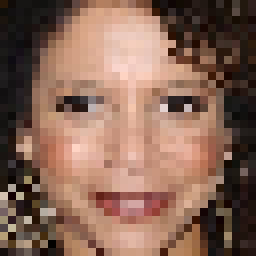}} \\[.1cm]

{\includegraphics[width=.29\linewidth]{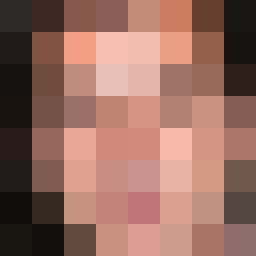}} &
{\includegraphics[width=.29\linewidth]{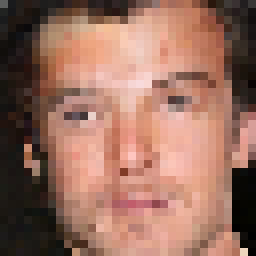}} &
{\includegraphics[width=.29\linewidth]{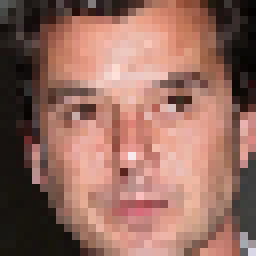}} \\[.1cm]

{\includegraphics[width=.29\linewidth]{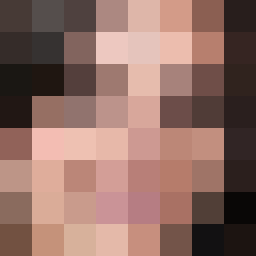}} &
{\includegraphics[width=.29\linewidth]{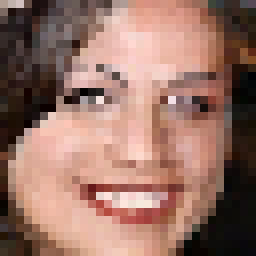}} &
{\includegraphics[width=.29\linewidth]{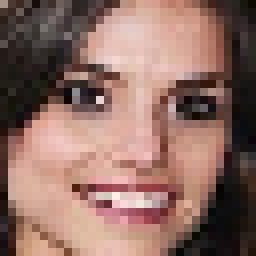}} \\
\end{tabular}
\end{center}
\caption{Illustration of our probabilistic \ourname super
  resolution model trained end-to-end on a dataset of celebrity
  faces. The left column shows $8\!\times\!8$ low resolution inputs
  from the test set. The middle and last columns show $32\!\times\!32$
  images as predicted by our model \vs the ground truth. Our
  model incorporates strong face priors to synthesize realistic hair
  and skin details.}
\label{fig:sr-fig1}
\end{figure}

This paper studies super resolution with particularly small inputs and
large magnification ratios, where the amount of information available
to accurately construct a high resolution image is very limited
(\figref{fig:sr-fig1}, left column). Thus, the problem is
underspecified and many plausible, high resolution images may match a
given low resolution input image. Building improved models for
state-of-the-art in super resolution in the high magnification regime
is significant for improving the state-of-art in super resolution, and
more generally for building better conditional generative models of
images \cite{PixelCNN, DCGAN, CONDITIONALGAN, PixelRNN}.

As the magnification ratio increases, a super resolution model need
not only account for textures, edges, and other low-level statistics
\cite{huang1999statistics, shan2008fast, kim2010single}, but must
increasingly account for complex variations of objects, viewpoints,
illumination, and occlusions. At increasing levels of magnification,
the details do not exist in the source image anymore, and the
predictive challenge shifts from recovering details (\eg~deconvolution
\cite{kundur1996blind}) to synthesizing plausible novel details {\it
  de novo} \cite{DCGAN, PixelCNN}.

%% Specifically, generating super
%% resolution images at high magnification requires making discretely selecting
%% and drawing textures, shapes and patterns at different parts of an
%% image.

% A great test bed for examining this problem are human faces.
% Recognizing faces has a rich history in the computer vision literature
% (see references in \cite{facereview, facenet}). Faces are particularly interesting because
% humans are especially sensitive to the shape and details of faces and building
% a system that sufficiently convince a human is challenging task.

Consider a low resolution image of a face in \figref{fig:sr-fig1}, left column.
In such $8\!\times\!8$ pixel images the fine spatial details
of the hair and the skin are missing and cannot be
faithfully restored with
interpolation techniques \cite{cubic_interpolation}.
However, by incorporating prior knowledge of
faces and their typical variations, a sketch artist might be able to
imagine and draw believable details using specialized software packages
\cite{laughery1980sketch}.

In this paper, we show how a fully {\em
  probabilistic} model that is trained {\em end-to-end} using a log-likelihood objective
can play the role of such an artist by synthesizing $32\!\times\!32$ face
images depicted in \figref{fig:sr-fig1}, middle column.
We find that drawing multiple samples from this model produces high resolution images that
exhibit multi-modality, resembling the diversity of images that plausibly
correspond to a low resolution image.
In human evaluation studies we demonstrate
that naive human observers can easily distinguish real images from the
outputs of sophisticated super resolution models using deep networks and
mean squared error (MSE) objectives \cite{VDSR}. However,
samples drawn from our probabilistic model are able fool a human
observer up to $27.9\%$ of the time -- compared to a chance rate of $50\%$.

In summary, the main contributions of the paper include:
\vspace*{-.55cm}
\begin{itemize}
  \setlength{\parskip}{0pt}
  \setlength{\itemsep}{3pt plus 1pt}
\item Characterization of the {\em underspecified} super resolution problem
  in terms of multi-modal prediction.
\item Proposal of a new probabilistic model tailored to the super
  resolution problem, which produces diverse, plausible non-blurry high
  resolution samples.
\item Proposal of a new loss term for conditional probabilistic models
  with powerful autoregressive decoders to avoid the conditioning
  signal to be ignored.
\item Human evaluation demonstrating that traditional metrics in super
  resolution (\eg~pSNR and SSIM) fail to capture sample quality in
  the regime of underspecified super resolution.
\end{itemize}
We proceed by describing related work, followed by explaining how the
multi-modal problem is not addressed using traditional
objectives. Then, we propose a new probabilistic model building on top
of ResNet~\cite{resnet16} and PixelCNN~\cite{PixelRNN}. The paper
highlights the diversity of high resolution samples generated by the
model and demonstrates the quality of the samples through human
evaluation studies.

\section{Related work}

Super resolution has a long history in computer vision~\cite{survey}.
% and many methods exist for approaching this problem.
Methods relying on interpolation
\cite{cubic_interpolation}
are easy to implement and widely used, however these methods suffer from
a lack of expressivity since linear models cannot express complex
dependencies between the inputs and outputs.
% reconstruct complex scenery.
In practice, such methods often fail to adequately predict
high frequency details leading to blurry
high resolution outputs.

Enhancing linear methods with rich image priors such as sparsity
\cite{Aharon} or Gaussian mixtures \cite{mixtures}
have substantially improved the
quality of the methods; likewise, leveraging low-level image statistics such as
edge gradients improves predictions
\cite{StructuredFaceHallucination, sun2008image, Fattal:2007:IUV:1276377.1276496, huang1999statistics,
shan2008fast, kim2010single}.
Much work has been done on algorithms that search a database of patches
and combine them to create plausible high frequency details in
zoomed images \cite{freeman2002example, Huang-CVPR-2015}. 
Recent patch-based work has focused on improving basic interpolation methods by
building a dictionary of pre-learned filters on images and selecting the
appropriate patches by an efficient hashing mechanism \cite{raisr}. Such
dictionary methods have improved the inference speed while
being comparable to state-of-the-art.

Another approach for super resolution is to abandon inference speed requirements
and focus on constructing the high resolution images at increasingly
higher magnification factors. Convolutional neural networks (CNNs) represent an approach to
the problem that avoids explicit dictionary construction, but rather
implicitly extracts multiple layers of abstractions by learning layers of filter kernels.
Dong~\etal~\cite{SRCNN} employed a three layer CNN with MSE
loss.
Kim~\etal~\cite{VDSR} improved accuracy by increasing the depth
to $20$ layers and learning only the residuals between the high
resolution image and an interpolated low resolution image. Most
recently, SRResNet~\cite{SRGAN} uses many ResNet blocks to achieve
state of the art pSNR and SSIM on standard super resolution
benchmarks--we employ a similar design for our conditional network and catchall
regression baseline.

%error gradients from deep networks can be used.

Instead of using a per-pixel loss, Johnson~\etal \cite{Johnson} use Euclidean distance between activations of
a pre-trained CNN for model's predictions \vs ground truth images.
Using this so-called preceptual loss, they train feed-forward networks for super resolution and style transfer. 
Bruna~\etal~\cite{Bruna} also use perceptual loss
to train a super resolution network, but inference is done via gradient propagation
to the low-res input (\eg~\cite{gatys}).

Another promising direction has been to employ an adversarial loss for training a network.
A super-resolution network is trained in opposition to a secondary network that attempts
to discriminate whether or not a synthesized high resolution image is real or fake.
Networks trained with traditional $L_p$ losses (e.g. \cite{VDSR, SRCNN}) suffer from blurry
images, where as networks employing an adversarial loss predict compelling, high frequency detail \cite{SRGAN, Yu2016}.
S{\o}nderby~\etal~\cite{2016arXiv161004490K} employed networks trained with
adversarial losses but constrained the network to learn affine transformations
that ensures the
model only generate images that downscale back to the low resolution
inputs. S{\o}nderby~\etal~\cite{2016arXiv161004490K} also explore a
masked autoregressive model but without the gated layers
and using a mixture of gaussians instead of a multinomial distribution.
Denton \etal~\cite{nips2015denton} use a multi-scale adversarial
network for image synthesis that is amenable for super-resolutions tasks.

Although generative adversarial networks (GANs)~\cite{GAN} provide a
promising direction, such networks suffer from several drawbacks:
first, training an adversarial network is unstable \cite{DCGAN} and
many methods are being developed to increase the robustness of
training \cite{unrolled_gans}.  Second, GANs suffer from a common
failure case of mode collapse \cite{unrolled_gans} where by the
resulting model produces samples that do not capture the diversity of
samples available in the training data. Finally, tracking the
performance of adversarial networks is challenging because it is
difficult to associate a probabilistic interpretation to their
results.  These points motivate approaching the problem with a
distinct approach to permit covering of the full diversity of the
training dataset.

PixelRNN and PixelCNN \cite{PixelRNN, PixelCNN} are
probabilistic generative models that impose an order on image pixels 
in order to represent them as a long sequence.
The probability of subsequent pixels is conditioned on previously observed pixels.
One variant of PixelCNN \cite{PixelCNN} obtained state-of-the-art predictive ability in terms of log-likelihood on academic benchmarks such as CIFAR-10 and MNIST.
Since PixelCNN uses log-likelihood for training, the model is penalized if
negligible probability is assigned to any of the training examples.
%PixelCNNs assign probability mass to every training example they see, which
%means they are theoretically able to model the entire dataset distribution.
By contrast, adversarial networks only learn enough to fool a
non-stationary discriminator.  This latter point suggests that a
PixelCNN might be able to predict a large diversity of high resolution
images that might be associated with a given low resolution
image. Further, using log-likelihood as the training objective allows
for hyper parameter search to find models within a model family by
simply comparing their log probabilities on a validation set.

\section{Probabilistic super resolution}
\label{sec:method}

We aim to learn a probabilistic super resolution model that discerns
the statistical dependencies between a high resolution image and a
corresponding low resolution image. Let $\bx$ and $\by$ denote a low
resolution and a high resolution image, and let $\bys$ represent a
ground-truth high resolution image. In order to learn a parametric
model of $\modelp(\by \mid \bx)$, we exploit a large dataset of pairs
of low resolution inputs and ground-truth high resolution outputs,
denoted $\dataset \equiv \{(\bx^{(i)}, \by^{*(i)})\}_{i=1}^N$. One can
easily collect such a large dataset by starting from some high
resolution images and lowering the resolution as much as needed. To
optimize the parameters $\btheta$ of the conditional distribution $p$,
we maximize a conditional log-likelihood objective defined as,
\begin{equation}
O(\btheta \mid \dataset) = \sum_{(\bx, \bys) \in \dataset}
\log p(\bys \mid \bx)~.
\label{eq:cll-objective}
\end{equation}
\comment{
Once a model is trained and appropriate parameters $\bthetah$ are
found, at inference time, one can draw samples from the model
$\modelph(\by \mid \bx)$.
}

The key problem discussed in this paper is the exact form of $p(\by
\mid \bx)$ that enables efficient learning and inference, while
generating realistic non-blurry outputs. We first discuss
pixel-independent models that assume that each output pixel is
generated with an independent stochastic process given the input. We
elaborate why these techniques result in sub-optimal blurry super
resolution results. Then, we describe our \ourname super resolution
model that generates output pixels one at a time to enable modeling
the statistical dependencies between the output pixels, resulting in
sharp synthesized images given very low resolution inputs.

%% , and synthesizes
%% sharp images from very blurry input.
%% using PixelCNN~\cite{PixelRNN,PixelCNN}
\subsection{Pixel independent super resolution}

The simplest form of a probabilistic super resolution model assumes
that the output pixels are conditionally independent given the
inputs. As such, the conditional distribution of $p(\by \mid \bx)$
factors into a product of independent pixel predictions. Suppose an
RGB output $\by$ has $M$ pixels each with three color channels,
\ie~$\by \in \Real^{3M}$. Then,
\begin{equation}
\log p(\by \mid \bx) = \sum_{i=1}^{3M} \,\log p(\y_{i} \mid \bx)~.
\end{equation}
Two general forms of pixel prediction models have been explored in the
literature: {\em Gaussian} and {\em multinomial} distributions to
model continuous and discrete pixel values respectively. In the
Gaussian case,
\begin{equation}
\log p(\y_{i} \mid \bx) = -\frac{1}{2\sigma^2}\,\lVert \y_{i} -
C_i(\bx)\rVert_2^2 - \log \sqrt{2\sigma^2\pi}~,
\label{eq:pixel-gaussian}
\end{equation}
where $C_i(\bx)$ denotes the $\th{i}$ element of a non-linear
transformation of $\bx$ via a convolutional neural network.
Accordingly, $C_i(\bx)$ is the estimated mean for the $\th{i}$ output
pixel $\y_{i}$, and $\sigma^2$ denotes the variance. Often the
variance is not learned, in which case maximizing the conditional
log-likelihood of \eqref{eq:cll-objective} reduces to minimizing the
MSE between $\y_{i}$ and $C_i(\bx)$ across the pixels and channels
throughout the dataset. Super resolution models based on MSE
regression fall within this family of pixel independent models
\cite{SRCNN, VDSR, SRGAN}.  Implicitly, the outputs of a neural
network parameterize a set of Gaussians with fixed variance.  It is
easy to verify that the joint distribution $p(\y \mid \bx)$ is
unimodal as it forms an isotropic multi-variate Gaussian.

\comment{Mean squared error regression can be interpreted as a
probabilistic model where the outputs of the network are
parameterizing fix-width Gaussians.}

Alternatively, one could discrete the output dimensions into $K$
possible values (\eg~$K=256$), and use a multinomial distribution as
the predictive model for each pixel~\cite{zhang2016colorful}, where
$\y_{i} \in \{1, \ldots, K\}$.  The pixel prediction model based on a
multinomial softmax operator is represented as,
\vspace*{-.2cm}
\begin{equation}
p(\y_{i} = k \mid \bx) = \frac{\exp\{ C_{ik}(\bx)\}}{\sum_{v=1}^K \exp\{ C_{iv}(\bx)\}}~,
\label{eq:pixel-softmax}
\end{equation}
where a network with a set of softmax weights,
$\{\bw_{jk}\}_{j=1,k=1}^{3,K}$, for each value per color channel is
used to induce $C_{ik}(\bx)$. Even though $p(\y_{i} \mid \bx)$ in
\eqref{eq:pixel-softmax} can express multimodal distributions, the
conditional dependency between the pixels cannot be captured, \ie~the
model cannot choose between drawing an edge at one position
\vs~another since that requires coordination between the samples.

%% %% coloring an apple red or
%% %% green each both modes exist for each pixel.

%That's all to say that MSE regression CNNs and pixel-independent cross-entropy
%CNNs are also fully probabilistic models. 

\comment{
Pixel independent super resolution models are considered the state of
the art as measured by image similarity metrics such as pSNR and
structural similarity (SSIM)~\cite{SSIM}, but they fail to produce
photo realistic images. In particular, the super resolution outputs
from these models often look blurry at the edge boundaries and lack
sufficient details, especially if the magnification ratio is large.
}

\subsection{Synthetic multimodal task}

\begin{figure}
\begin{center}
{\bf How the dataset was created}
\vspace{0.2cm}\\
\includegraphics[width=.25\textwidth]{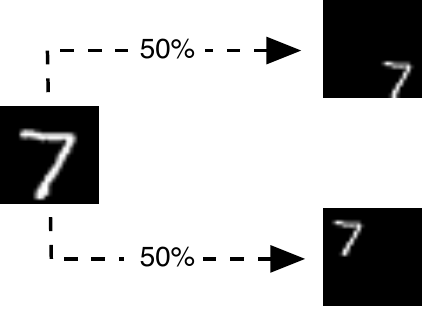}
\vspace{0.3cm}\\
{
\begin{tabular}{rc@{\hspace{.1cm}}c@{\hspace{.1cm}}c@{\hspace{.1cm}}c@{\hspace{.1cm}}}

\multicolumn{5}{c}{\bf Samples from trained model \vspace{.1cm}}\\

 \raisebox{.5cm}{\small $L_2$ regression}  & {\includegraphics[width=.14\linewidth]{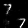}} &{\includegraphics[width=.14\linewidth]{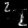}} &{\includegraphics[width=.14\linewidth]{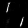}} &{\includegraphics[width=.14\linewidth]{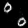}} \\ 
 \raisebox{.5cm}{\small cross-entropy}  & {\includegraphics[width=.14\linewidth]{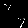}} &{\includegraphics[width=.14\linewidth]{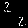}} &{\includegraphics[width=.14\linewidth]{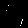}} &{\includegraphics[width=.14\linewidth]{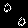}} \\ 
 \raisebox{.5cm}{\small PixelCNN}  & {\includegraphics[width=.14\linewidth]{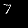}} &{\includegraphics[width=.14\linewidth]{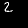}} &{\includegraphics[width=.14\linewidth]{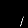}} &{\includegraphics[width=.14\linewidth]{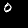}} \\[-.4cm]

\end{tabular}
}
\end{center}

\caption{Simulated dataset demonstrates challenge of multimodal prediction.
{\it Top}: Synthesized dataset in which samples are randomly translated to top-left or bottom-right corners. 
{\it Bottom}: Example predictions for various algorithms trained on this
  dataset. The pixel independent $L_2$ regression and
  cross-entropy models fail to predict a single mode but instead predict a blend of two spatial locations even though such samples do not exist in the training set.
Conversely, the PixelCNN stochastically predicts the location of the digit at either corner with mutual exclusion.}
\label{fig:toy}
\end{figure}

To demonstrate how pixel independent models fail at conditional image modeling,
we create a synthetic dataset that explicitly requires multimodal prediction.
For many
dense image predictions tasks, e.g. super resolution \cite{survey},
colorization \cite{zhang2016colorful, diversecolorization2016}, and
depth estimation \cite{Saxena05learningdepth},
models that are able to predict a single mode are heavily preferred
over models that blend modes together.
For example, in the task of colorization
selecting a strong red or green for an apple is better than selecting a
brown-toned color that reflects the smeared average of all of the apple
colors observed in the training set.

We construct a simple multimodal {\it MNIST corners} dataset to demonstrate
the challenge of this problem. {\it MNIST corners} is constructed by randomly
placing an MNIST digit in either the top-left or bottom-right corner (\figref{fig:toy}, top). Several networks are trained to predict individual
samples from this dataset to demonstrate the unique challenge of this simple
example.

The challenge behind this toy example is for a network to exclusively predict
an individual digit in a corner of an image. Training a moderate-sized
10-layer
convolutional neural network ($\sim100K$ parameters) with an $L_2$ objective
(i.e. MSE regression) results in blurry image samples in which the two modes are blended
together (\figref{fig:toy}, {\it $L_2$ regression}). That is, {\it never} in the dataset 
does an example image contain a digit 
in both corners, yet this model incorrectly predicts a blend of such samples.
Replacing the loss with a discrete, per-pixel cross-entropy produces
sharper images but likewise fails to stochastically predict a digit in a corner
of the image (\figref{fig:toy}, {\it cross-entropy}).

%In this synthetic task, the input is an MNIST digit
%($1^{\text{st}}$ row of \figref{fig:toy}), and the output is the same
%input digit but scaled and translated either into the upper left
%corner or upper right corner ($2^{\text{nd}}$ and $3^{\text{rd}}$ rows
%of \figref{fig:toy}). The dataset has an equal ratio of upper left and
%upper right outputs, which we call the {\it MNIST corners} dataset.

%A convolutional network using per pixel squared error loss (\figref{fig:toy},
%$L_2$ Regression) produces two blurry figures. Replacing the continuous
%loss with a per-pixel cross-entropy produces
%crisper images but also fails to capture the stochastic bimodality because
%both digits are shown in both corners (\figref{fig:toy}, cross-entropy).

\section{\Ourname super resolution}
The lack of conditional independence between predicted pixels is a significant
failure mode for the previous probabilistic objectives in the synthetic example
(Equations \ref{eq:pixel-gaussian} and \ref{eq:pixel-softmax}).
%There are two general methods to model statistical
%correlations between output pixels. 
One approach to this problem is to define the
conditional distribution of the output pixels jointly as a
multivariate Gaussian mixture~\cite{zoran2011} or an undirected
graphical model~\cite{freeman2000markov}. Both of these conditional
distributions require constructing a statistical dependency
between output pixels for which inference may be computationally
expensive.

A second approach is to
factorize the joint distribution using the chain rule
by imposing an order on image pixels,
\begin{equation}
\log p(\by \mid \bx) = \sum_{i=1}^{M} \,\log p(\by_{i} \mid \bx,
\by_{<i})~,
\label{eq:autoregressive-logp}
\end{equation}
where the generation of each output dimension is conditioned on the
input and previous output pixels \cite{larochelle2011, NIPS2013_5060}.
We denote the conditioning\footnote{Note that in color images one must impose an order on both spatial
locations as well as color channels. In a color image the conditioning
is based on the the input and previously outputted pixels at previous spatial locations as well as pixels at the same spatial location.}
up to pixel $i$ by $\by_{<i}$ where $\{\by_1,\ldots,\by_{i-1}\}$.
The benefits of this approach
%, called {\em \ourname} super resolution, 
are that the exact form of the
conditional dependencies is flexible and the inference is
straightforward.

PixelCNN is a stochastic model that provides
an explicit model for $\log p(\by_{i} \,|\,\bx, \by_{<i})~$ as a gated,
hierarchical chain of cleverly masked convolutions \cite{PixelRNN, PixelCNN, PixelCNNpp}.
The goal of PixelCNN
is to capture multi-modality and capture pixel correlations in an image. Indeed,
training a PixelCNN on the {\it MNIST corners} dataset successfully
captures the bimodality of the
problem and produces sample in which digits reside exclusively in a single
corner (\figref{fig:toy}, {\it PixelCNN}).
Importantly, the model never predicts both digits simultaneously.

Applying the PixelCNN to a super-resolution problem is a straightforward
application that requires modifying the architecture to supply a conditioning
on a low resolution version of the image.
In early experiments we found the auto-regressive distribution of the model
largely ignore the conditioning of the low resolution image. This phenomenon
referred to as ``optimization challenges'' has been readily documented in
the context of sequential autoencoder models \cite{bowman2015} (see also
\cite{serban, NIPS2016_6275} for more discussion).

%Inspired by the PixelCNN model, we use a multinomial
%distribution to model discrete pixel values in Eq.
%\ref{eq:autoregressive-logp}. Alternatively, one could use an
%autoregressive prediction model with Gaussian or Logistic (mixture)
%conditionals as proposed in~\cite{PixelCNNpp}.

\begin{figure}[t]
\begin{center}
\includegraphics[width=.85\linewidth]{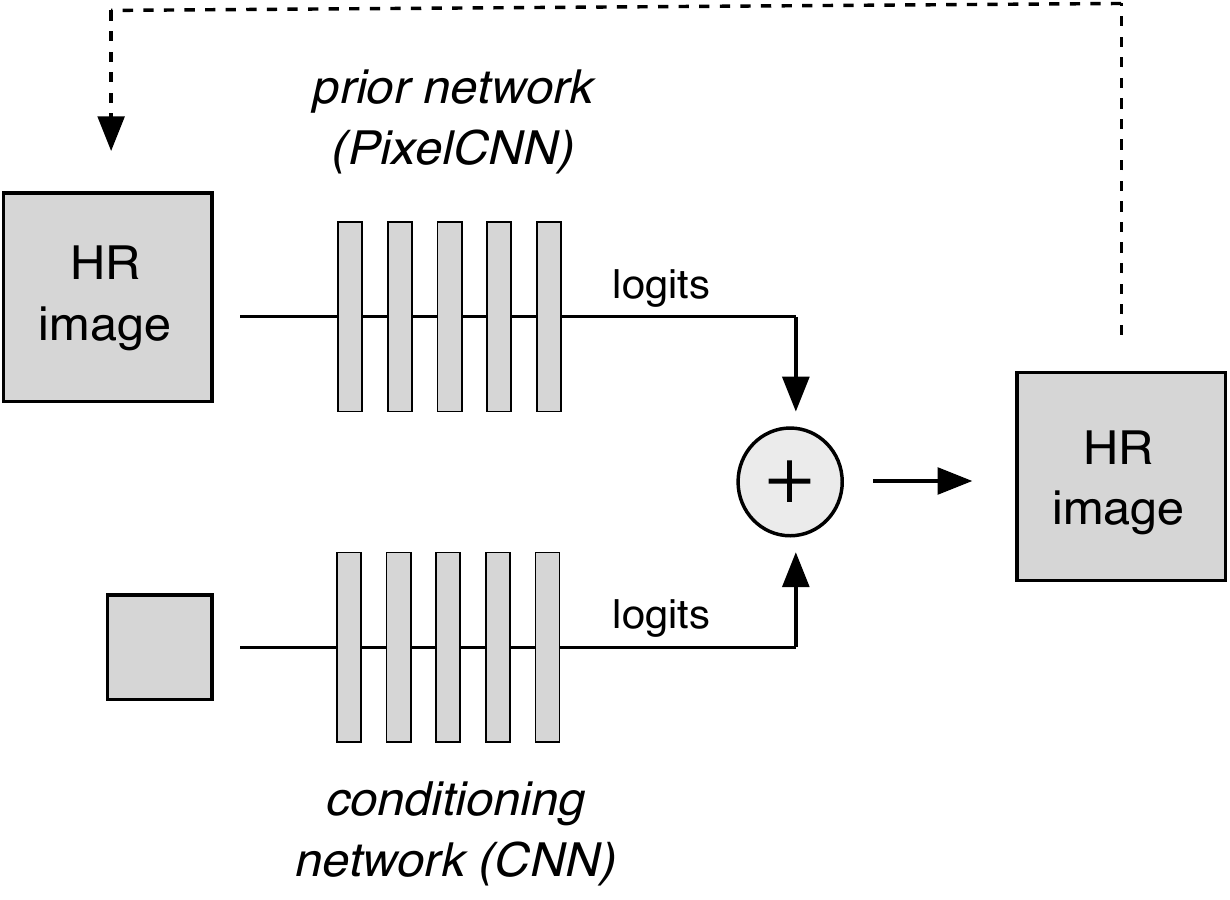}
\vspace*{-.2cm}
\end{center}
\caption{The proposed super resolution network comprises a
  \textit{conditioning network} and a \textit{prior network}.  The
  \textit{conditioning network} is a CNN that receives a low
  resolution image as input and outputs logits predicting the
  conditional log-probability of each high resolution (HR) image
  pixel. The \textit{prior network}, a PixelCNN
  \cite{PixelCNN}, makes predictions based on previous stochastic
  predictions (indicated by dashed line). The model's probability
  distribution is computed as a softmax operator on top of the sum of
  the two sets of logits from the prior and conditioning networks.}
\label{fig:model-components}
\end{figure}

To address this issue we modify the architecture of PixelCNN
to more explicitly depend on the conditioning of a low resolution image.
In particular, we propose a late fusion model \cite{KarpathyCVPR14} that 
factors the problem into auto-regressive and conditioning
components (Figure \ref{fig:model-components}).
The auto-regressive portion of the model, termed a {\it
prior network} captures the serial dependencies of
the pixels while the conditioning component, termed a {\it conditioning network}
captures the global structure of the low resolution image. Specifically,
we formulate the prior network to be a PixelCNN and the conditioning network
to be a deep convolutional network employed previously for super resolution~\cite{SRGAN}.

%The model, outlined in \figref{fig:model-components}, comprises two
%major components that are fused together at a late stage and trained
%jointly: (1) a {\em conditioning} network and (2) a {\em prior}
%network.

%The conditioning network is a pixel independent prediction
%model that maps a low resolution image to a probabilistic skeleton of
%a high resolution image, while the prior network is supposed to add
%natural high resolution details to make the outputs look more
%realistic.

Given an input $\bx \in \Real^L$, let $\cond_i(\bx): \Real^L \to
\Real^K$ denote a conditioning network predicting a vector of logit
values corresponding to the $K$ possible values that the $\th{i}$
output pixel can take. Similarly, let $\prior_i(\by_{< i}):
\Real^{i-1} \to \Real^K$ denote a prior network predicting a vector of
logit values for the $\th{i}$ output pixel. Our probabilistic model
predicts a distribution over the $\th{i}$ output pixel by simply
adding the two sets of logits and applying a softmax operator on them,
\begin{equation}
  p(y_i \mid \bx, \by_{<i}) = \softmax(\cond_i(\bx) + \prior_i(\by_{< i}))~.
\label{eq:sr-model}
\end{equation}

To optimize the parameters of $\cond$ and $\prior$ jointly, we perform
stochastic gradient ascent to maximize the conditional log likelihood
in~\eqref{eq:cll-objective}. That is, we optimize a cross-entropy
loss between the model's predictions in \eqref{eq:sr-model} and
discrete ground truth labels $\ys_i \in \{1, \ldots, K\}$,
\begin{equation}
\begin{aligned}
O_1 = \!\!\sum_{(\bx, \bys) \in \dataset} \sum_{i=1}^M
~&\Big(\trans{\1{\bys_i}} \!\left( \cond_i(\bx) + \prior_i(\bys_{<i}) \right)\\ 
& ~~~-
\logsumexp( \cond_i(\bx) + \prior_i(\bys_{<i}) )\Big)~,
\end{aligned}
\label{eq:obj-asr1}
\end{equation}
where $\logsumexp(\cdot)$ is the log-sum-exp operator corresponding to
the log of the denominator of a softmax, and $\1{k}$ denotes a
$K$-dimensional one-hot indicator vector with its $\th{k}$ dimension set to $1$.

Our preliminary experiments indicate that models trained with
\eqref{eq:obj-asr1} tend to ignore the conditioning network as the
statistical correlation between a pixel and previous high resolution
pixels is stronger than its correlation with low resolution inputs. To
mitigate this issue, we include an additional loss in our objective to
enforce the conditioning network to be optimized. This additional loss
measures the cross-entropy between the conditioning network's
predictions via $\softmax(\cond_i(\bx))$ and ground truth labels. The
total loss that is optimized in our experiments is a sum of two
cross-entropy losses formulated as,
\begin{equation}
\begin{aligned}
O_2 = \!\!&\sum_{(\bx, \bys) \in \dataset} \sum_{i=1}^M
~\Big(\trans{\1{\bys_i}} \!\left( 2\,\cond_i(\bx) + \prior_i(\bys_{<i}) \right)\\ 
&~~~~
- \logsumexp( \cond_i(\bx) + \prior_i(\bys_{<i}) ) - \logsumexp( \cond_i(\bx) )\Big)~.
\end{aligned}
\label{eq:obj-asr2}
\end{equation}

\comment{
To train we use two cross-entropy objectives to jointly train the model. One
to constrain the conditioning network to produce high resolution images,
$O_c$, and the other to train the PixelCNN with using the mixture of the two
likelihood densities, $O_p$.
\begin{equation}
O_c = \sum_{i=1}^M \bys_i \cdot A_i(x)
%  O &=& O_c + O_p
\end{equation}

Where $i$ is a sub-pixel index, and $A_i(x)$ is the $i$th output of the
conditioning network before the softmax, $B_i(\bys)$ is the $i$th output of the
prior network, and $\bys_i$ is a one hot vector representing the ground truth
sub-pixel at location $i$.  
}

Once the network is trained, sampling from the model is
straightforward. Using \eqref{eq:sr-model}, starting at $i=1$, first
we sample a high resolution pixel. Then, we proceed pixel by pixel,
feeding in the previously sampled pixel values back into the network, and
draw new high resolution pixels. The three channels of each pixel are
generated sequentially in turn.

We additionally consider {\em greedy decoding}, where one
always selects the pixel value with the largest probability and
sampling from a tempered softmax, where the concentration of a
distribution $p$ is adjusted by using a temperature parameter $\tau > 0$,
\begin{equation*}
p_\tau = \frac{ p^{(1/\tau)} }{ \lVert p^{(1/\tau)} \rVert_1 }~.
\end{equation*}
To control the concentration of our sampling distribution $p(\y_i \mid
\bx,\by_{<i})$, it suffices to divide the logits from $\cond$ and
$\prior$ by a parameter $\tau$. Note that as $\tau$ goes towards
$0$, the distribution converges to the mode. 

%% \footnote{We use a
%%   non-standard notion of temperature that represents $\frac{1}{\tau}$
%%   in the standard notation.}, and sampling converges to greedy
%% decoding

\subsection{Implementation details \label{sec:details}}
We summarize the network architecture for the pixel recursive
super resolution model.
The conditioning architecture is similar in design to SRResNet~\cite{SRGAN}.
The conditioning network is a feed-forward convolutional neural
network that takes a low resolution image through a series of
$18-30$ ResNet blocks \cite{resnet16} and
transposed convolution layers \cite{odena2016deconvolution}.
The last layer uses a
$1\!\times\!1$ convolution to increase the number of channels to
predict a multinomial distribution over 
$256$ possible color channel
values for each sub-pixel.
%$the channels to
%$256\!\times\!3$ and uses the resulting activations to predict a
%multinomial distribution over $256$ possible sub-pixel values via a
%softmax operator.
%\comment{ This conditioning network is trained with
%  a cross-entropy loss against the true high resolution image.}
%This network provides the ability to absorb the global structure of
%the image in the marginal probability distribution of the pixels.
%\todo{Link to b256 image in table?}
%Due to the softmax layer it can capture the rich intricacies of the
%high resolution distribution, but we have no way to coherently sample
%from it.  Sampling sub-pixels independently will mix the assortment of
%distributions.
The prior network architecture consists of 20 gated PixelCNN blocks 
with 32 channels at each layer \cite{PixelCNN}. The final layer 
of the super-resolution network is a softmax operation over the sum
of the activations from the conditioning and prior networks.
%The prior network provides a way to tie together the sub-pixel
%distributions and allow us to take samples dependent on each other.
%We use 20 gated PixelCNN layers with 32 channels at each layer.
%We leave conditioning until the late stages of the network, where
%we add the pre-softmax activations from the conditioning network and
%prior network before computing the final joint softmax distribution.
The model is built by using TensorFlow~\cite{tf} and trained across
$8$ GPUs with synchronous SGD updates. For training details and a
complete list of architecture parameters, please see
\ifthenelse{\boolean{arxiv}}{Appendix~\ref{hparams}}{\supp{}}.

\section{Experiments}

We assess the effectiveness of the proposed \ourname super resolution
method on two datasets containing centrally cropped faces (CelebA
\cite{liu2015faceattributes}) and bedroom images (LSUN Bedrooms
\cite{lsun}). In both datasets we resize the images to $8\!\times\!8$
and $32\!\times\!32$ pixels with bicubic interpolation to provide the
input $\bx$ and output $\by$ for training and evaluation.

We compare our technique against three baselines including (1)
\textbf{\NN{}}; a nearest neighbor search baseline inspired by
previous work on example-based super
resolution~\cite{freeman2002example}, (2) \boldregression{}; a deep
neural network using Resnet blocks trained with MSE objective, and (3)
\textbf{\srez}; a GAN based super resolution model implemented
by~\cite{srez} similar to \cite{Yu2016}. We exclude the results of the
\srez baseline on bedrooms dataset as they are not competitive, and the
model was developed specifically for faces.

The \NN{} baseline computes $\by$ for a sample $\bx$ by searching the
training set $\dataset = \{(\bx^{(i)}, \by^{*(i)})\}_{i=1}^N$ for the
nearest example indexed by $i^* = {\operatorname{argmin}}_{i} \Vert
\bx^{(i)} - \bx\rVert_2^2$, and returns the high resolution
counterpart $\by^{*(i^*)}$.  The \NN{} baseline is a representative
result of exemplar based super resolution approaches, and helps us
test whether the model performs a naive lookup from the training
dataset.

The \regression{} baseline employs a design similar to
SRResNet~\cite{SRGAN} that reports state-of-the-art in terms of image
similarity metrics%
\footnote{ Note that the regression architecture is nearly identical
  to the conditioning network in Section \ref{sec:details}. The slight
  change is to force the network to predict bounded values in RGB
  space. To enforce this behavior, the top layer is outputs three
  channels instead of one and employ a $\tanh(\cdot)$ instead of a
  ReLU$(\cdot)$ nonlinearity.}. Most significantly, we alter the
network to compute the residuals with respect to a bicubic
interpolation of the input \cite{VDSR}.  The $L_2$ regression provides
a comparison to a state-of-the-art convolutional network that performs
a unimodal pixel independent prediction.

The \srez{} super resolution baseline~\cite{srez} exploits a
conditional GAN architecture, and combines an adversarial loss with a
{\em consistency} loss, which encourages the low-resolution version of
predicted $\by$ to be close to $\bx$ as measures by $L_1$. There is a
weighting between the two losses specified by~\cite{srez} as $0.9$ for
the consistency and $0.1$ for the adversarial loss, and we keep them
the same in our face experiments.

\subsection{Super resolution samples}

% Insert the bedroom figure.
\begin{figure}
\begin{center}
\def\s{.27}
\begin{tabular}{@{\hspace{.2cm}}c@{\hspace{.2cm}}c@{\hspace{.2cm}}c@{\hspace{.2cm}}}
%\begin{tabular}{@{\hspace{.1cm}}c@{\hspace{.1cm}}c@{\hspace{.1cm}}c@{\hspace{.1cm}}}
$8\!\times\!8$ input & $32\!\times\!32$ samples & ground truth\\[.2cm]
%Input & 32 & G. Truth \\[.2cm]
%\begin{tabular}{@{\hspace{.1cm}}c@{\hspace{.1cm}}c@{\hspace{.1cm}}c@{\hspace{.1cm}}c@{\hspace{.1cm}}c@{\hspace{.1cm}}}
%Input & Regression & Ours & G. Truth & NN \\[.2cm]
{\includegraphics[width=\s\linewidth]{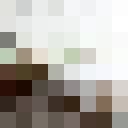}} &
%{\includegraphics[width=.18\linewidth]{lsun_bedroom_samples6/baselineL2/0911bd_128px.png}} &
{\includegraphics[width=\s\linewidth]{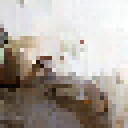}} &
{\includegraphics[width=\s\linewidth]{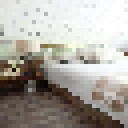}} \\
%{\includegraphics[width=.18\linewidth]{lsun_bedroom_samples6/nearest_neighbors_L2/0911bd.png}} \\

%{\includegraphics[width=\s\linewidth]{lsun_bedroom_samples6/lowres/f07c62_128px.png}} &
%{\includegraphics[width=.18\linewidth]{lsun_bedroom_samples6/baselineL2/f07c62_128px.png}} &
%{\includegraphics[width=\s\linewidth]{lsun_bedroom_samples6/mask120/f07c62_128px.png}} &
%{\includegraphics[width=\s\linewidth]{lsun_bedroom_samples6/truth/f07c62_128px.png}} \\
%{\includegraphics[width=.18\linewidth]{lsun_bedroom_samples6/nearest_neighbors_L2/f07c62.png}} \\

{\includegraphics[width=\s\linewidth]{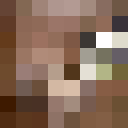}} &
%{\includegraphics[width=.18\linewidth]{lsun_bedroom_samples6/baselineL2/79473d_128px.png}} &
{\includegraphics[width=\s\linewidth]{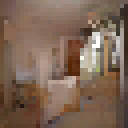}} &
{\includegraphics[width=\s\linewidth]{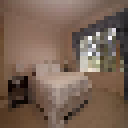}} \\
%{\includegraphics[width=.18\linewidth]{lsun_bedroom_samples6/nearest_neighbors_L2/79473d.png}}

{\includegraphics[width=\s\linewidth]{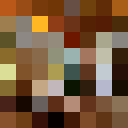}} &
%{\includegraphics[width=.18\linewidth]{lsun_bedroom_samples6/baselineL2/38ba86_128px.png}} &
{\includegraphics[width=\s\linewidth]{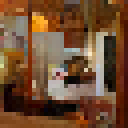}} &
{\includegraphics[width=\s\linewidth]{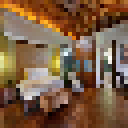}} \\[-.4cm]
%{\includegraphics[width=.18\linewidth]{lsun_bedroom_samples6/nearest_neighbors_L2/38ba86.png}} 
 
% {\includegraphics[width=.18\linewidth]{lsun_bedroom_samples6/lowres/92a2d1_128px.png}} &
% {\includegraphics[width=.18\linewidth]{lsun_bedroom_samples6/baselineL2/92a2d1_128px.png}} &
% {\includegraphics[width=.18\linewidth]{lsun_bedroom_samples6/mask120/92a2d1_128px.png}} &
% {\includegraphics[width=.18\linewidth]{lsun_bedroom_samples6/truth/92a2d1_128px.png}} &
% {\includegraphics[width=.18\linewidth]{lsun_bedroom_samples6/nearest_neighbors_L2/92a2d1.png}} \\
 
% {\includegraphics[width=.18\linewidth]{lsun_bedroom_samples6/lowres/2abb8f_128px.png}} &
% {\includegraphics[width=.18\linewidth]{lsun_bedroom_samples6/baselineL2/2abb8f_128px.png}} &
% {\includegraphics[width=.18\linewidth]{lsun_bedroom_samples6/mask120/2abb8f_128px.png}} &
% {\includegraphics[width=.18\linewidth]{lsun_bedroom_samples6/truth/2abb8f_128px.png}} &
% {\includegraphics[width=.18\linewidth]{lsun_bedroom_samples6/nearest_neighbors_L2/2abb8f.png}} \\

%{\includegraphics[width=.18\linewidth]{lsun_bedroom_samples6/lowres/c6de71_128px.png}} &
%{\includegraphics[width=.18\linewidth]{lsun_bedroom_samples6/baselineL2/c6de71_128px.png}} &
%{\includegraphics[width=.18\linewidth]{lsun_bedroom_samples6/mask120/c6de71_128px.png}} &
%{\includegraphics[width=.18\linewidth]{lsun_bedroom_samples6/truth/c6de71_128px.png}} &
%{\includegraphics[width=.18\linewidth]{lsun_bedroom_samples6/nearest_neighbors_L2/c6de71.png}} \\[-.4cm]

\end{tabular}
\end{center}
\caption{Illustration of our probabilistic \ourname super
  resolution model trained end-to-end on LSUN Bedrooms dataset.}
\label{fig:bedrooms}
\end{figure}
 \begin{figure}
\def\gw{.19}
\begin{center}
\begin{tabular}{c@{\hspace{.2cm}}c@{\hspace{.1cm}}c@{\hspace{.1cm}}c@{\hspace{.1cm}}c}
% /cns/od-d/home/rld/super_resolution/exp102/A_one_hot/samples/
% /cns/od-d/home/rld/super_resolution/exp148/mask_bedroom_b10/samples/1.30/
\subfloat{\includegraphics[width=\gw\linewidth]{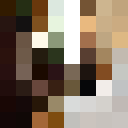}} &
\subfloat{\includegraphics[width=\gw\linewidth]{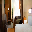}} &
\subfloat{\includegraphics[width=\gw\linewidth]{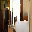}} &
\subfloat{\includegraphics[width=\gw\linewidth]{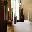}} &
\subfloat{\includegraphics[width=\gw\linewidth]{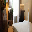}} \\[-.4cm]

\subfloat{\includegraphics[width=\gw\linewidth]{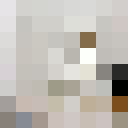}} &
\subfloat{\includegraphics[width=\gw\linewidth]{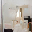}} &
\subfloat{\includegraphics[width=\gw\linewidth]{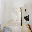}} &
\subfloat{\includegraphics[width=\gw\linewidth]{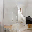}} &
\subfloat{\includegraphics[width=\gw\linewidth]{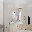}} \\[-.4cm]

\subfloat{\includegraphics[width=\gw\linewidth]{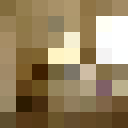}} &
\subfloat{\includegraphics[width=\gw\linewidth]{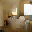}} &
\subfloat{\includegraphics[width=\gw\linewidth]{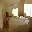}} &
\subfloat{\includegraphics[width=\gw\linewidth]{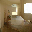}} &
\subfloat{\includegraphics[width=\gw\linewidth]{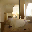}} \\[-.4cm]

\subfloat{\includegraphics[width=\gw\linewidth]{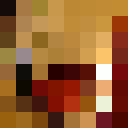}} &
\subfloat{\includegraphics[width=\gw\linewidth]{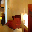}} &
\subfloat{\includegraphics[width=\gw\linewidth]{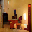}} &
\subfloat{\includegraphics[width=\gw\linewidth]{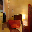}} &
\subfloat{\includegraphics[width=\gw\linewidth]{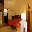}} \\[-.4cm]

\subfloat{\includegraphics[width=\gw\linewidth]{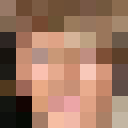}} &
\subfloat{\includegraphics[width=\gw\linewidth]{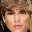}} &
\subfloat{\includegraphics[width=\gw\linewidth]{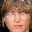}} &
\subfloat{\includegraphics[width=\gw\linewidth]{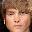}} &
\subfloat{\includegraphics[width=\gw\linewidth]{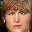}} \\[-.4cm]

\subfloat{\includegraphics[width=\gw\linewidth]{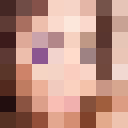}} &
\subfloat{\includegraphics[width=\gw\linewidth]{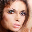}} &
\subfloat{\includegraphics[width=\gw\linewidth]{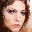}} &
\subfloat{\includegraphics[width=\gw\linewidth]{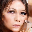}} &
\subfloat{\includegraphics[width=\gw\linewidth]{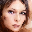}} \\[-.4cm]

\subfloat{\includegraphics[width=\gw\linewidth]{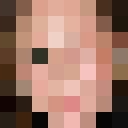}} &
\subfloat{\includegraphics[width=\gw\linewidth]{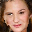}} &
\subfloat{\includegraphics[width=\gw\linewidth]{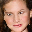}} &
\subfloat{\includegraphics[width=\gw\linewidth]{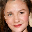}} &
\subfloat{\includegraphics[width=\gw\linewidth]{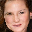}} \\[-.4cm]

\subfloat{\includegraphics[width=\gw\linewidth]{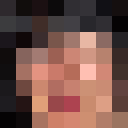}} &
\subfloat{\includegraphics[width=\gw\linewidth]{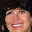}} &
\subfloat{\includegraphics[width=\gw\linewidth]{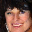}} &
\subfloat{\includegraphics[width=\gw\linewidth]{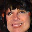}} &
\subfloat{\includegraphics[width=\gw\linewidth]{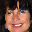}} \\[-.4cm]

\end{tabular}
\end{center}
\caption{Diversity of samples from pixel recursive super resolution model. Left column: Low resolution input. Right columns: Multiple super resolution samples at $\tau = 0.8$ conditioned upon low resolution input.}
\label{fig:multimodal}
\vspace*{-.4cm}
\end{figure}
 High
resolution samples generated by the pixel recursive super resolution
capture the rich structure of the dataset and appear perceptually
plausible (Figure \ref{fig:sr-fig1} and \ref{fig:bedrooms};
\ifthenelse{\boolean{arxiv}}{Appendix~\ref{sec:sup_bedrooms} and
  \ref{sec:sup_faces}}{\supp{}}).  Sampling from the super resolution
model multiple times results in different high resolution images for a
given low resolution image (Figure
\ref{fig:multimodal}\ifthenelse{\boolean{arxiv}}{;
  Appendix~\ref{sec:sup_bedrooms} and \ref{sec:sup_faces}}{ and
  \supp{}}).  Qualitatively, the samples from the model identify many
plausible high resolution images with distinct qualitative features
that correspond to a given lower resolution image.  Note that the
differences between samples for the faces dataset are far less drastic
than seen in our synthetic dataset, where failure to cleanly predict
modes indicated complete failure.

\begin{figure*}
\def\s{.0915}
\def\ss{.15cm}
\begin{center}
\begin{small}
\begin{tabular}{@{\hspace{\ss}}c@{\hspace{\ss}}c@{\hspace{\ss}}c@{\hspace{\ss}}c@{\hspace{\ss}}c@{\hspace{\ss}}c@{\hspace{\ss}}c@{\hspace{\ss}}c@{\hspace{\ss}}c@{\hspace{\ss}}c}
\multicolumn{8}{c}{}\\[.1cm]
Input & G. Truth & \NN{} & GAN~\cite{srez} & Bicubic & \regression & Greedy & $\tau=1.0$ & $\tau=0.9$ & $\tau=0.8$ \\
{\includegraphics[width=\s\linewidth]{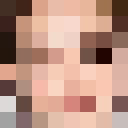}} &
{\includegraphics[width=\s\linewidth]{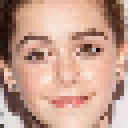}} &
{\includegraphics[width=\s\linewidth]{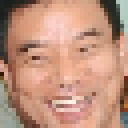}} &
{\includegraphics[width=\s\linewidth]{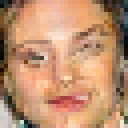}} &
{\includegraphics[width=\s\linewidth]{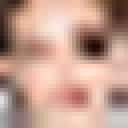}} &
{\includegraphics[width=\s\linewidth]{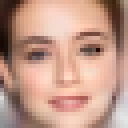}} &
{\includegraphics[width=\s\linewidth]{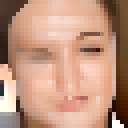}} &
{\includegraphics[width=\s\linewidth]{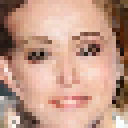}} &
{\includegraphics[width=\s\linewidth]{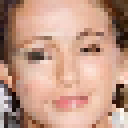}} &
{\includegraphics[width=\s\linewidth]{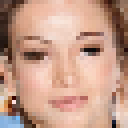}} \\
-- & 2.85  & 2.74  & -- & 1.76  & 2.34  & 1.82  & 2.94  & 2.79  & 2.69 \\ % 24bd27

 {\includegraphics[width=\s\linewidth]{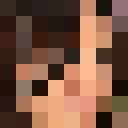}} &
 {\includegraphics[width=\s\linewidth]{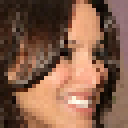}} &
 {\includegraphics[width=\s\linewidth]{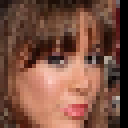}} &
 {\includegraphics[width=\s\linewidth]{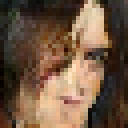}} &
 {\includegraphics[width=\s\linewidth]{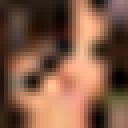}} &
 {\includegraphics[width=\s\linewidth]{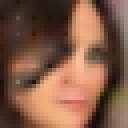}} &
 {\includegraphics[width=\s\linewidth]{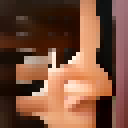}} &
 {\includegraphics[width=\s\linewidth]{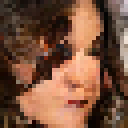}} &
 {\includegraphics[width=\s\linewidth]{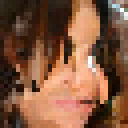}} &
 {\includegraphics[width=\s\linewidth]{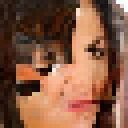}} \\
-- & 2.96  & 2.71  & --   & 1.82 & 2.17  & 1.77  & 3.18  & 3.09  & 2.95 \\ % 1b8b00
 
{\includegraphics[width=\s\linewidth]{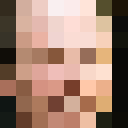}} &
{\includegraphics[width=\s\linewidth]{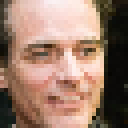}} &
{\includegraphics[width=\s\linewidth]{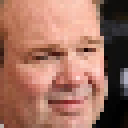}} &
{\includegraphics[width=\s\linewidth]{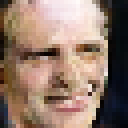}} &
{\includegraphics[width=\s\linewidth]{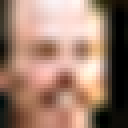}} &
{\includegraphics[width=\s\linewidth]{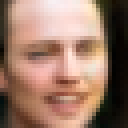}} &
{\includegraphics[width=\s\linewidth]{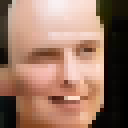}} &
{\includegraphics[width=\s\linewidth]{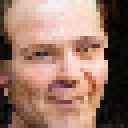}} &
{\includegraphics[width=\s\linewidth]{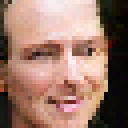}} &
{\includegraphics[width=\s\linewidth]{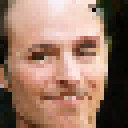}} \\
-- & 2.76  & 2.63  & -- & 1.80  & 2.35  & 1.64  & 2.99  & 2.90  & 2.64 \\[-.4cm] % 15ef4f

\end{tabular}
\end{small}
\end{center}
\caption{Comparison of super resolution models. Columns from left to
  right include input, Ground truth, \NN{} (nearest neighbor super
  resolution), \srez{}, bicubic upsampling, \regression (neural
  network optimized with MSE), greedy decoding is pixel recursive
  model, followed by sampling with various temperatures ($\tau$)
  controlling the concentration of the predictive
  distribution. Negative log-probabilities are reported below the
  images. Note that the best log-probability is associated with
  bicubic upsampling and greedy decoding even though the images are
  poor quality.}
\label{fig:nll}
\end{figure*}

The quality of samples is sensitive to the temperature
(\figref{fig:nll}, right columns). Greedy decoding ($\tau=0$) results
in poor quality samples that are overly smooth and contain horizontal
and vertical line artifacts.  Samples from the default temperature
($\tau=1.0$) are perceptually more plausible, although they tend to
contain undesired high frequency content. Tuning the temperature
($\tau$) between $0.9$ and $0.8$ proves beneficial for improving the
quality of the samples.

\subsection{Quantitative evaluation of image similarity}

Many methods exist for quantifying image similarity that attempt to measure
human perception judgements of similarity \cite{SSIM, MS-SSIM, MAD}. We quantified
the prediction accuracy of our model compared to ground truth
using pSNR and MS-SSIM (Table~\ref{tab:results}). We found that
our own visual assessment of the predicted image quality did not correspond to these
image similarities metrics. For instance, bicubic interpolation achieved relatively
high metrics even though the samples appeared quite poor. This result matches
recent observations that suggest that pSNR and SSIM provide poor judgements
of super resolution quality when new details are synthesized  \cite{SRGAN, Johnson}. In addition, Figure~\ref{fig:nll} highlights how the perceptual quality of model samples
do not necessarily correspond to negative log likelihood (NLL). Smaller NLL means the model has assigned
that image a larger probability mass. The greedy, bicubic, and regression faces
are preferred by the model despite exhibiting worse perceptual quality.

%To check that samples do indeed correspond to the low-resolution input, we
%measured the $L_2$ between the input and downsampled samples (Table~\ref{tab:results}).
%Although our samples are less consistent than the baseline model, they are
%much better than the nearest neighbor telling us that the samples do indeed
%correspond to the low-resolution version.

We next measured how well the high resolution samples corresponded to the low resolution input by measuring the consistency. The consistency is quantified as $L_2$ distance between the low-resolution input image and a bicubic downsampled version of the high resolution estimate. Lower consistencies indicate superior correspondence with the low-resolution image.  Note that this is an explicit objective the GAN~\cite{srez}. The pixel recursive model achieved consistencies on par with the $L_2$ regression model and bicubic interpolation indicating that even though the model was producing diverse samples, the samples were largely constrained by the low-resolution image. Most importantly, the pixel recursive model achieved superior consistencies then the GAN~\cite{srez} even though the model does not explicitly optimize for this criterion.%
\footnote{Note that one may improve the consistency of the GAN by increasing its weight in the objective. Increasing the weight for the consistency term will likely lead to decreased perceptual quality in the images but improved consistency. Regardless, the images generated by the pixel recursive model are superior in both consistency and perceptual quality as judged humans for a range of temperatures.}

The consistency measure additionally provided an important control experiment to determine if the pixel recursive model were just naively copying the nearest training sample. If the pixel recursive model were just copying the nearest training sample, then the consistency of the \NN{} model would be equivalent to the pixel recursive model. We instead find that the pixel recursive model has superior consistency values indicating that the model is not just naively copying the closest training examples.

%To ensure that samples do indeed correspond to the low-resolution input, we measured how consistent the high resolution output image is with the low resolution input image (Table~\ref{tab:results}, 'consistency'). Specifically, we measured the L2 distance between the low-resolution input image and a bicubic downsampled version of the high resolution estimate. Lower L2 distances correspond to high resolutions that are more similar to the original low resolution image. Note that the nearest neighbor high resolution images are less consistent even though we used a database of 3 million training images to search for neighbors in the case of LSUN bedrooms. In contrast, the bicubic resampling and the PixelCNN upsampling methods showed consistently better consistency with the low resolution image. This indicates that our samples do indeed correspond to the low-resolution input.

% Created with //google3/experimental/users/rld/calculate_algo_metrics.py
\def\sss{.2cm}
\begin{table}
\begin{center}
\begin{small}
\begin{tabular}{@{}r@{\hspace{\sss}}|@{\hspace{\sss}}c@{\hspace{\sss}}c@{\hspace{\sss}}c@{\hspace{\sss}}c@{\hspace{\sss}}p{1.4cm}@{}}
%\begin{tabular}{p{1.4cm}|p{0.6cm}p{0.6cm}p{1.35cm}p{1.2cm}p{1.5cm}}
{\it CelebA}  & pSNR & SSIM  & MS-SSIM & Consistency & \% Fooled\\
%{\footnotesize {\it CelebA}}  & {\footnotesize pSNR} & {\footnotesize SSIM}  & {\footnotesize MS-SSIM} & {\footnotesize Consistency} & {\footnotesize \% Fooled}\\
\hline
\hline 
 Bicubic    & 28.92 & 0.84 & 0.76             &  0.006   & -- \\
 \NN{}         & 28.18 & 0.73 & 0.66             &  0.024      & -- \\
 \regression & \bf 29.16 & \bf 0.90 & \bf 0.90 &  \bf 0.004         & $4.0\pm0.2$  \\
 GAN~\cite{srez} & 28.19 & 0.72 &  0.67 &  0.029         & $8.5\pm0.2$ \\
 $\tau=1.0$ & 29.09 & 0.84 & 0.86             &  0.008        & $\bf 11.0\pm0.1$ \\
 $\tau=0.9$ & 29.08 & 0.84 & 0.85             &  0.008        & $10.4\pm0.2$  \\
 $\tau=0.8$ & 29.08 & 0.84 & 0.86             &  0.008        & $10.2\pm0.1$ \\
\end{tabular}

\vspace{0.2cm}
\begin{tabular}{@{}r@{\hspace{\sss}}|@{\hspace{\sss}}c@{\hspace{\sss}}c@{\hspace{\sss}}c@{\hspace{\sss}}c@{\hspace{\sss}}p{1.4cm}@{}}
{\it LSUN}  & pSNR & SSIM  & MS-SSIM & Consistency & \% Fooled\\
\hline
\hline 
 Bicubic    & \bf 28.94 & 0.70 & 0.70      & \bf  0.002    & -- \\
 \NN{}         & 28.15 & 0.49 & 0.45          & 0.040    & -- \\
 \regression & 28.87 & \bf 0.74 & \bf 0.75  & 0.003     & $2.1\pm0.1$      \\
 $\tau=1.0$ & 28.92 & 0.58 & 0.60          & 0.016    & $17.7\pm0.4$     \\
 $\tau=0.9$ & 28.92 & 0.59 & 0.59          & 0.017   & $22.4\pm0.3$     \\
 $\tau=0.8$ & 28.93 & 0.59 & 0.58          & 0.018   & $\bf 27.9\pm0.3$     \\
\end{tabular}
\end{small}
\end{center}
% faces bicubic
% MSE truth 0.000
% MSE bicubic 0.006
% MSE nearest_neighbors_L2 0.024
% MSE baselineL2 0.004
% MSE mask100 0.008
% MSE mask110 0.008
% MSE mask120 0.008
% 
% bedrooms bicubic
% MSE truth 0.014
% MSE bicubic 0.002
% MSE nearest_neighbors_L2 0.040
% MSE baselineL2 0.003
% MSE mask100 0.016
% MSE mask110 0.017
% MSE mask120 0.018

\vspace*{-.3cm}
\caption{Test results on the cropped CelebA (top) and LSUN Bedroom
  (bottom) datasets magnified from $8\!\times\!8$ to
  \mbox{$32\!\times\!32$}. We report pSNR, SSIM, and MS-SSIM between
  samples and the ground truth.  Consistency measures the MSE between
  the low-resolution input and a corresponding downsampled output.  \%
  Fooled reports measures how often the algorithms' outputs fool a
  human in a crowd sourced study; 50\% would be perfectly confused.}
\label{tab:results}
\vspace*{-.3cm}
\end{table}

\subsection{Perceptual evaluation with humans}
\label{subsec:human}

Given that automated quantitative measures did not match our
perceptual judgements, we conducted a human study to assess the
effectiveness of the super resolution algorithm.  In particular, we
performed a forced choice experiment on crowd-sourced workers in order
to determine how plausible a given high resolution sample is from each
model. Following \cite{zhang2016colorful},
each worker was presented a true image and a corresponding prediction
from a model, and asked ``{\it Which image, would you guess, is from a
  camera?}''. We performed this study across 283 workers on Amazon
Mechanical Turk and statistics were accrued across 40 unique workers
for each super resolution algorithm.\footnote{Specifically, each
  worker was given one second to make a forced choice
  decision. Workers began a session with 10 practice questions during
  which they received feedback. The practice pairs were not counted in
  the results. After the practice pairs, each worker was shown 45
  additional pairs.  A subset of the pairs were simple, {\it golden}
  questions designed to constantly check if the worker was paying
  attention. Data from workers that answered golden questions
  incorrectly were thrown out.}
%We continued running sessions until forty different
%different workers were tested on each of the four algorithms. 

%The golden question pits a bicubicly upsampled image (very blurry)
%against the ground truth.
%Excluding the golden and practice questions, we
%count forty answers per session.  Sessions in which they missed any golden
%questions are thrown out. Workers were only allowed to participate in any of
%our studies once.  We continued running sessions until forty different
%different workers were tested on each of the four algorithms. 

Table~\ref{tab:results} reports the percentage of samples for a given algorithm
that a human incorrectly believed to be a real image. Note that a perfect
algorithm would fool a human at rate of 50\%. The $L_2$ regression model
fooled humans 2-4\% of the time and the GAN~\cite{srez} fooled humans 8.5\% of
the time. The pixel recursive model fooled humans 11.0\%
and 27.9\% of the time for faces and bedrooms, respectively -- significantly
above the state-of-the-art regression model. Importantly, we found that the
selection of the sampling temperature $\tau$ greatly influenced
the quality of the samples and in turn the fraction of time that humans were 
fooled. 
Nevertheless the pixel recursive model outperformed the strongest baseline
model, the GAN, across all temperatures. 
A ranked list of the best and worst fooling examples is reproduced in 
\ifthenelse{\boolean{arxiv}}{Appendix~\ref{sec:human_ratings}}{the \supp{}}
along with the fool rates.

\section{Conclusion}

We advocate research on super resolution with high magnification
ratios, where the problem is dramatically underspecified as high
frequency details are missing. Any model that produces non-blurry super
resolution outputs must make sensible predictions of the missing
content to operate in such a heavily multimodal regime. We present a
fully probabilistic method that tackles super resolution with small
inputs, demonstrating that even $8\!\times\!8$ images can be enlarged
to sharp $32\!\times\!32$ images. Our technique outperforms several
strong baselines including the ones optimizing a regression objective
or an adversarial loss.  We perform human evaluation studies showing
that samples from the \ourname model look more plausible to humans,
and more generally, common metrics like pSNR and SSIM do not correlate
with human judgment when the magnification ratio is large.

%% We introduced a dataset with a small number of explicit modes
%% to demonstrate the failure cases of two common pixel independent
%% likelihood models.

\section*{Acknowledgments}

We thank A{\"a}ron van den Oord, Sander Dieleman, and the Google Brain team for
insightful comments and discussions.

{\small
\bibliographystyle{ieee}
\bibliography{paper}

\begin{thebibliography}{10}\itemsep=-1pt

\bibitem{tf}
M.~Abadi, A.~Agarwal, P.~Barham, E.~Brevdo, Z.~Chen, C.~Citro, G.~S. Corrado,
  A.~Davis, J.~Dean, M.~Devin, S.~Ghemawat, I.~Goodfellow, A.~Harp, G.~Irving,
  M.~Isard, Y.~Jia, R.~Jozefowicz, L.~Kaiser, M.~Kudlur, J.~Levenberg,
  D.~Man\'{e}, R.~Monga, S.~Moore, D.~Murray, C.~Olah, M.~Schuster, J.~Shlens,
  B.~Steiner, I.~Sutskever, K.~Talwar, P.~Tucker, V.~Vanhoucke, V.~Vasudevan,
  F.~Vi\'{e}gas, O.~Vinyals, P.~Warden, M.~Wattenberg, M.~Wicke, Y.~Yu, and
  X.~Zheng.
\newblock {TensorFlow}: Large-scale machine learning on heterogeneous systems,
  2015.
\newblock Software available from tensorflow.org.

\bibitem{Aharon}
M.~Aharon, M.~Elad, and A.~Bruckstein.
\newblock Svdd: An algorithm for designing overcomplete dictionaries for sparse
  representation.
\newblock {\em Trans. Sig. Proc.}, 54(11):4311--4322, Nov. 2006.

\bibitem{bowman2015}
S.~R. Bowman, L.~Vilnis, O.~Vinyals, A.~M. Dai, R.~J{\'{o}}zefowicz, and
  S.~Bengio.
\newblock Generating sentences from a continuous space.
\newblock {\em CoRR}, abs/1511.06349, 2015.

\bibitem{Bruna}
J.~Bruna, P.~Sprechmann, and Y.~LeCun.
\newblock Super-resolution with deep convolutional sufficient statistics.
\newblock {\em CoRR}, abs/1511.05666, 2015.

\bibitem{nips2015denton}
E.~L. Denton, S.~Chintala, A.~Szlam, and R.~Fergus.
\newblock Deep generative image models using a laplacian pyramid of adversarial
  networks.
\newblock {\em NIPS}, 2015.

\bibitem{diversecolorization2016}
A.~Deshpande, J.~Lu, M.~Yeh, and D.~A. Forsyth.
\newblock Learning diverse image colorization.
\newblock {\em CoRR}, abs/1612.01958, 2016.

\bibitem{SRCNN}
C.~Dong, C.~C. Loy, K.~He, and X.~Tang.
\newblock Image super-resolution using deep convolutional networks.
\newblock {\em CoRR}, abs/1501.00092, 2015.

\bibitem{Fattal:2007:IUV:1276377.1276496}
R.~Fattal.
\newblock Image upsampling via imposed edge statistics.
\newblock {\em ACM Trans. Graph.}, 26(3), July 2007.

\bibitem{freeman2002example}
W.~T. Freeman, T.~R. Jones, and E.~C. Pasztor.
\newblock Example-based super-resolution.
\newblock {\em IEEE Computer graphics and Applications}, 2002.

\bibitem{freeman2000markov}
W.~T. Freeman and E.~C. Pasztor.
\newblock Markov networks for super-resolution.
\newblock In {\em CISS}, 2000.

\bibitem{srez}
D.~Garcia.
\newblock srez: Adversarial super resolution.
\newblock \url{https://github.com/david-gpu/srez}, 2016.

\bibitem{gatys}
L.~A. Gatys, A.~S. Ecker, and M.~Bethge.
\newblock A neural algorithm of artistic style.
\newblock {\em CoRR}, abs/1508.06576, 2015.

\bibitem{GAN}
I.~Goodfellow, J.~Pouget-Abadie, M.~Mirza, B.~Xu, D.~Warde-Farley, S.~Ozair,
  A.~Courville, and Y.~Bengio.
\newblock Generative adversarial nets, 2014.

\bibitem{resnet16}
K.~He, X.~Zhang, S.~Ren, and J.~Sun.
\newblock Deep residual learning for image recognition.
\newblock {\em CVPR}, 2015.

\bibitem{cubic_interpolation}
H.~Hou and H.~Andrews.
\newblock {Cubic splines for image interpolation and digital filtering}.
\newblock {\em Acoustics, Speech and Signal Processing, IEEE Transactions on},
  26(6):508--517, Jan. 2003.

\bibitem{huang1999statistics}
J.~Huang and D.~Mumford.
\newblock Statistics of natural images and models.
\newblock In {\em Computer Vision and Pattern Recognition, 1999. IEEE Computer
  Society Conference on.}, volume~1. IEEE, 1999.

\bibitem{Huang-CVPR-2015}
J.-B. Huang, A.~Singh, and N.~Ahuja.
\newblock Single image super-resolution from transformed self-exemplars.
\newblock In {\em IEEE Conference on Computer Vision and Pattern Recognition)},
  2015.

\bibitem{Johnson}
J.~Johnson, A.~Alahi, and F.~Li.
\newblock Perceptual losses for real-time style transfer and super-resolution.
\newblock {\em CoRR}, abs/1603.08155, 2016.

\bibitem{2016arXiv161004490K}
C.~{Kaae S{\o}nderby}, J.~{Caballero}, L.~{Theis}, W.~{Shi}, and
  F.~{Husz{\'a}r}.
\newblock {Amortised MAP Inference for Image Super-resolution}.
\newblock {\em ArXiv e-prints}, Oct. 2016.

\bibitem{KarpathyCVPR14}
A.~Karpathy, G.~Toderici, S.~Shetty, T.~Leung, R.~Sukthankar, and L.~Fei-Fei.
\newblock Large-scale video classification with convolutional neural networks.
\newblock In {\em CVPR}, 2014.

\bibitem{VDSR}
J.~Kim, J.~K. Lee, and K.~M. Lee.
\newblock Accurate image super-resolution using very deep convolutional
  networks.
\newblock {\em CoRR}, abs/1511.04587, 2015.

\bibitem{kim2010single}
K.~I. Kim and Y.~Kwon.
\newblock Single-image super-resolution using sparse regression and natural
  image prior.
\newblock {\em IEEE Transactions on Pattern Analysis and Machine Intelligence},
  32(6):1127--1133, 2010.

\bibitem{kundur1996blind}
D.~Kundur and D.~Hatzinakos.
\newblock Blind image deconvolution.
\newblock {\em IEEE signal processing magazine}, 13(3):43--64, 1996.

\bibitem{larochelle2011}
H.~Larochelle and I.~Murray.
\newblock The neural autoregressive distribution estimator.
\newblock In {\em The Proceedings of the 14th International Conference on
  Artificial Intelligence and Statistics}, volume~15 of {\em JMLR: W\&CP},
  pages 29--37, 2011.

\bibitem{laughery1980sketch}
K.~R. Laughery and R.~H. Fowler.
\newblock Sketch artist and identi-kit procedures for recalling faces.
\newblock {\em Journal of Applied Psychology}, 65(3):307, 1980.

\bibitem{SRGAN}
C.~Ledig, L.~Theis, F.~Huszar, J.~Caballero, A.~Aitken, A.~Tejani, J.~Totz,
  Z.~Wang, and W.~Shi.
\newblock Photo-realistic single image super-resolution using a generative
  adversarial network.
\newblock {\em arXiv:1609.04802}, 2016.

\bibitem{liu2015faceattributes}
Z.~Liu, P.~Luo, X.~Wang, and X.~Tang.
\newblock Deep learning face attributes in the wild.
\newblock In {\em Proceedings of International Conference on Computer Vision
  (ICCV)}, 2015.

\bibitem{MAD}
K.~Ma, Q.~Wu, Z.~Wang, Z.~Duanmu, H.~Yong, H.~Li, and L.~Zhang.
\newblock Group mad competition - a new methodology to compare objective image
  quality models.
\newblock In {\em The IEEE Conference on Computer Vision and Pattern
  Recognition (CVPR)}, June 2016.

\bibitem{unrolled_gans}
L.~Metz, B.~Poole, D.~Pfau, and J.~Sohl{-}Dickstein.
\newblock Unrolled generative adversarial networks.
\newblock {\em CoRR}, abs/1611.02163, 2016.

\bibitem{CONDITIONALGAN}
M.~Mirza and S.~Osindero.
\newblock Conditional generative adversarial nets.
\newblock {\em CoRR}, abs/1411.1784, 2014.

\bibitem{survey}
K.~Nasrollahi and T.~B. Moeslund.
\newblock Super-resolution: A comprehensive survey.
\newblock {\em Mach. Vision Appl.}, 25(6):1423--1468, Aug. 2014.

\bibitem{odena2016deconvolution}
A.~Odena, V.~Dumoulin, and C.~Olah.
\newblock Deconvolution and checkerboard artifacts.
\newblock {\em Distill}, 2016.
\newblock http://distill.pub/2016/deconv-checkerboard.

\bibitem{DCGAN}
A.~Radford, L.~Metz, and S.~Chintala.
\newblock Unsupervised representation learning with deep convolutional
  generative adversarial networks.
\newblock {\em CoRR}, abs/1511.06434, 2015.

\bibitem{raisr}
Y.~Romano, J.~Isidoro, and P.~Milanfar.
\newblock {RAISR:} rapid and accurate image super resolution.
\newblock {\em CoRR}, abs/1606.01299, 2016.

\bibitem{FoE}
S.~Roth and M.~J. Black.
\newblock Fields of experts: A framework for learning image priors.
\newblock {\em CVPR}, 2005.

\bibitem{PixelCNNpp}
T.~Salimans, A.~Karpathy, X.~Chen, D.~P. Kingma, and Y.~Bulatov.
\newblock Pixelcnn++: A pixelcnn implementation with discretized logistic
  mixture likelihood and other modifications.
\newblock under review at ICLR 2017.

\bibitem{Saxena05learningdepth}
A.~Saxena, S.~H. Chung, and A.~Y. Ng.
\newblock Learning depth from single monocular images.
\newblock In {\em In NIPS 18}. MIT Press, 2005.

\bibitem{serban}
I.~V. Serban, A.~Sordoni, R.~Lowe, L.~Charlin, J.~Pineau, A.~C. Courville, and
  Y.~Bengio.
\newblock A hierarchical latent variable encoder-decoder model for generating
  dialogues.
\newblock {\em CoRR}, abs/1605.06069, 2016.

\bibitem{shan2008fast}
Q.~Shan, Z.~Li, J.~Jia, and C.-K. Tang.
\newblock Fast image/video upsampling.
\newblock {\em ACM Transactions on Graphics (TOG)}, 27(5):153, 2008.

\bibitem{NIPS2016_6275}
C.~K. S\o~nderby, T.~Raiko, L.~Maal\o~e, S.~r.~K. S\o~nderby, and O.~Winther.
\newblock Ladder variational autoencoders.
\newblock In D.~D. Lee, M.~Sugiyama, U.~V. Luxburg, I.~Guyon, and R.~Garnett,
  editors, {\em Advances in Neural Information Processing Systems 29}, pages
  3738--3746. Curran Associates, Inc., 2016.

\bibitem{sun2008image}
J.~Sun, Z.~Xu, and H.-Y. Shum.
\newblock Image super-resolution using gradient profile prior.
\newblock In {\em Computer Vision and Pattern Recognition, 2008. CVPR 2008.
  IEEE Conference on}, pages 1--8. IEEE, 2008.

\bibitem{NIPS2013_5060}
B.~Uria, I.~Murray, and H.~Larochelle.
\newblock Rnade: The real-valued neural autoregressive density-estimator.
\newblock In C.~J.~C. Burges, L.~Bottou, M.~Welling, Z.~Ghahramani, and K.~Q.
  Weinberger, editors, {\em Advances in Neural Information Processing Systems
  26}, pages 2175--2183. Curran Associates, Inc., 2013.

\bibitem{PixelRNN}
A.~van~den Oord, N.~Kalchbrenner, and K.~Kavukcuoglu.
\newblock Pixel recurrent neural networks.
\newblock {\em ICML}, 2016.

\bibitem{PixelCNN}
A.~van~den Oord, N.~Kalchbrenner, O.~Vinyals, L.~Espeholt, A.~Graves, and
  K.~Kavukcuoglu.
\newblock Conditional image generation with pixelcnn decoders.
\newblock {\em NIPS}, 2016.

\bibitem{SSIM}
Z.~Wang, A.~C. Bovik, H.~R. Sheikh, and E.~P. Simoncelli.
\newblock Image quality assessment: from error visibility to structural
  similarity.
\newblock {\em IEEE transactions on image processing}, 13(4):600--612, 2004.

\bibitem{MS-SSIM}
Z.~Wang, E.~P. Simoncelli, and A.~C. Bovik.
\newblock Multiscale structural similarity for image quality assessment.
\newblock In {\em Signals, Systems and Computers, 2004. Conference Record of
  the Thirty-Seventh Asilomar Conference on}, volume~2, pages 1398--1402. Ieee,
  2004.

\bibitem{StructuredFaceHallucination}
C.~Y. Yang, S.~Liu, and M.~H. Yang.
\newblock Structured face hallucination.
\newblock In {\em 2013 IEEE Conference on Computer Vision and Pattern
  Recognition}, pages 1099--1106, June 2013.

\bibitem{lsun}
F.~Yu, Y.~Zhang, S.~Song, A.~Seff, and J.~Xiao.
\newblock Lsun: Construction of a large-scale image dataset using deep learning
  with humans in the loop.
\newblock {\em arXiv preprint arXiv:1506.03365}, 2015.

\bibitem{Yu2016}
X.~Yu and F.~Porikli.
\newblock {\em Ultra-Resolving Face Images by Discriminative Generative
  Networks}, pages 318--333.
\newblock Springer International Publishing, Cham, 2016.

\bibitem{zhang2016colorful}
R.~Zhang, P.~Isola, and A.~A. Efros.
\newblock Colorful image colorization.
\newblock {\em ECCV}, 2016.

\bibitem{mixtures}
D.~Zoran and Y.~Weiss.
\newblock From learning models of natural image patches to whole image
  restoration.
\newblock In {\em Proceedings of the 2011 International Conference on Computer
  Vision}, ICCV '11, pages 479--486, Washington, DC, USA, 2011. IEEE Computer
  Society.

\bibitem{zoran2011}
D.~Zoran and Y.~Weiss.
\newblock From learning models of natural image patches to whole image
  restoration.
\newblock In {\em CVPR}, 2011.

\end{thebibliography}


\begin{thebibliography}{1}\itemsep=-1pt

\bibitem{srez}
D.~Garcia.
\newblock srez: Adversarial super resolution.
\newblock 2016.

\end{thebibliography}
}

\ifthenelse{\boolean{arxiv}}{
\newpage
\onecolumn
\appendix
% For submissions, this is included in supplemental.tex
% For arxiv this is included in paper.tex.

\section{Hyperparameters for pixel recursive super resolution model.} \label{hparams}
\begin{table}[h]
\centering
\begin{tabular}{@{}rlll@{}} \toprule
Operation               & Kernel       & Strides      & Feature maps  \\ \midrule
Conditional network -- $8 \times 8 \times 3$ input                        \\
$B$ $\times$ ResNet block            & $3 \times 3$          & 1          & $32$  \\
Transposed Convolution  & $3 \times 3$ & 2 & $32$  \\
$B$ $\times$ ResNet block            & $3 \times 3$          & 1          & $32$  \\
Transposed Convolution  & $3 \times 3$ & 2 & $32$  \\
$B$ $\times$ ResNet block            & $3 \times 3$          & 1          & $32$  \\
Convolution  & $1 \times 1$ & 1 & $3 * 256$  \\
\hline
PixelCNN network -- $32 \times 32 \times 3$ input                              \\
Masked Convolution  & $7 \times 7$ & 1 & $64$  \\
20 $\times$ Gated Convolution Blocks  & $5 \times 5$ & 1 & $64$  \\
Masked Convolution  & $1 \times 1$ & 1 & $1024$  \\
Masked Convolution  & $1 \times 1$ & 1 & $3 * 256$  \\
\hline
Optimizer     & \multicolumn{3}{@{}l@{}}{RMSProp (decay=0.95, momentum=0.9, epsilon=1e-8)}  \\
Batch size    & \multicolumn{3}{@{}l@{}}{32} \\
Iterations    & \multicolumn{3}{@{}l@{}}{2,000,000 for Bedrooms, 200,000 for faces.} \\
Learning Rate & \multicolumn{3}{@{}l@{}}{0.0004 and divide by 2 every 500000 steps.} \\
Weight, bias initialization  & \multicolumn{3}{@{}l@{}}{truncated normal (stddev=0.1), Constant($0$)} \\ \bottomrule
\end{tabular}
\vspace{0.2cm}
\caption{Hyperparameters used for both datasets. For LSUN bedrooms $B=10$ and for
the cropped CelebA faces $B=6$.}
\end{table}

\newpage
\section{Samples from models trained on LSUN bedrooms} \label{sec:sup_bedrooms}
\begin{longtable}{@{\hspace{.12cm}}c@{\hspace{.12cm}}c@{\hspace{.12cm}}c@{\hspace{.12cm}}c@{\hspace{.12cm}}c@{\hspace{.12cm}}c@{\hspace{.12cm}}c@{\hspace{.12cm}}c} 
Input & Bicubic & \regression & $\tau=1.0$ & $\tau=0.9$ & $\tau=0.8$ & Truth & \NN{} \\ 
 \endhead 
 {\includegraphics[width=.1\linewidth]{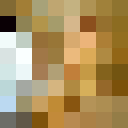}} 
 & {\includegraphics[width=.1\linewidth]{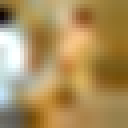}} 
 & {\includegraphics[width=.1\linewidth]{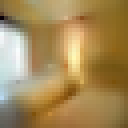}} 
 & {\includegraphics[width=.1\linewidth]{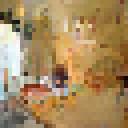}} 
 & {\includegraphics[width=.1\linewidth]{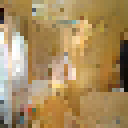}} 
 & {\includegraphics[width=.1\linewidth]{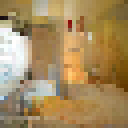}} 
 & {\includegraphics[width=.1\linewidth]{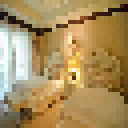}} 
 & {\includegraphics[width=.1\linewidth]{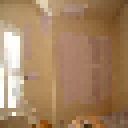}} 
 \\ 
{\includegraphics[width=.1\linewidth]{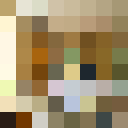}} 
 & {\includegraphics[width=.1\linewidth]{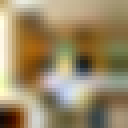}} 
 & {\includegraphics[width=.1\linewidth]{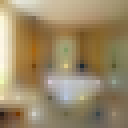}} 
 & {\includegraphics[width=.1\linewidth]{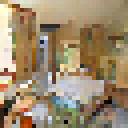}} 
 & {\includegraphics[width=.1\linewidth]{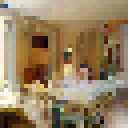}} 
 & {\includegraphics[width=.1\linewidth]{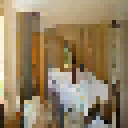}} 
 & {\includegraphics[width=.1\linewidth]{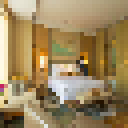}} 
 & {\includegraphics[width=.1\linewidth]{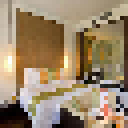}} 
 \\
\end{longtable}

\newpage
\section{Samples from models trained on CelebA faces} \label{sec:sup_faces}
\begin{longtable}{@{\hspace{.12cm}}c@{\hspace{.12cm}}c@{\hspace{.12cm}}c@{\hspace{.12cm}}c@{\hspace{.12cm}}c@{\hspace{.12cm}}c@{\hspace{.12cm}}c@{\hspace{.12cm}}c@{\hspace{.12cm}}c} 
Input & Bicubic & \regression & $\tau=1.0$ & $\tau=0.9$ & $\tau=0.8$ & Truth & \NN{} & GAN~\cite{srez} \\ 
 \endhead 
 {\includegraphics[width=.1\linewidth]{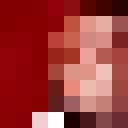}} 
 & {\includegraphics[width=.1\linewidth]{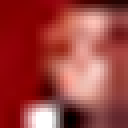}} 
 & {\includegraphics[width=.1\linewidth]{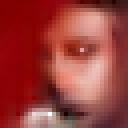}} 
 & {\includegraphics[width=.1\linewidth]{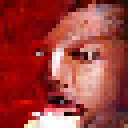}} 
 & {\includegraphics[width=.1\linewidth]{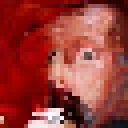}} 
 & {\includegraphics[width=.1\linewidth]{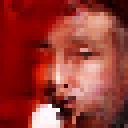}} 
 & {\includegraphics[width=.1\linewidth]{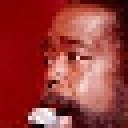}} 
 & {\includegraphics[width=.1\linewidth]{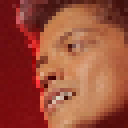}} 
 & {\includegraphics[width=.1\linewidth]{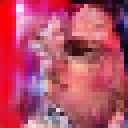}} 
 \\ 
{\includegraphics[width=.1\linewidth]{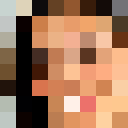}} 
 & {\includegraphics[width=.1\linewidth]{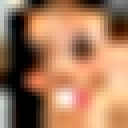}} 
 & {\includegraphics[width=.1\linewidth]{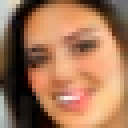}} 
 & {\includegraphics[width=.1\linewidth]{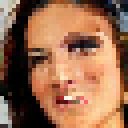}} 
 & {\includegraphics[width=.1\linewidth]{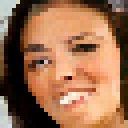}} 
 & {\includegraphics[width=.1\linewidth]{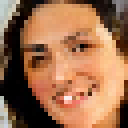}} 
 & {\includegraphics[width=.1\linewidth]{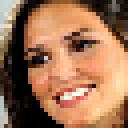}} 
 & {\includegraphics[width=.1\linewidth]{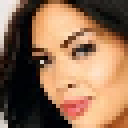}} 
 & {\includegraphics[width=.1\linewidth]{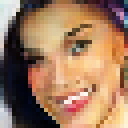}} 
 \\
\end{longtable}

\newpage
\section{Samples images that performed best and worst in human ratings.} \label{sec:human_ratings}

The best and worst rated images in the human study. The fractions below the images denote how many times a person choose that image over the ground truth. 

%\label{fig:bestworst}
\begin{figure*}[h]
\begin{center}
%%
\end{center}
%\caption{The best and worst rated images in the human study. The fractions below the images denote how many times a person choose that image over the ground truth. See the supplementary material for more images used in the study.}
%\label{fig:bestworst}
\end{figure*}

}{}

\end{document}